# Performance evaluation of parallel manipulators for milling application

A. Pashkevich[1,2], A. Klimchik[1,2], S. Briot[1], D. Chablat[1]
[1]Institut de Recherches en Communications et Cybernetique de Nantes, 1 rue de la No, 44321 Nantes, France
[2]Ecole des Mines de Nantes, 4 rue Alfred-Kastler, Nantes 44307, France
Anatol.Pashkevich@emn.fr, Alexandr.Klimchik@emn.fr, Sebastien.Briot@irccyn.ec-nantes.fr, Damien.Chablat@irccyn.ec-nantes.fr

**Abstract**
This paper focuses on the performance evaluation of the parallel manipulators for milling of composite materials. For this application the most significant performance measurements, which denote the ability of the manipulator for the machining are defined. In this case, optimal synthesis task is solved as a multicriterion optimization problem with respect to the geometric, kinematic, kinetostatic, elastostostatic, dynamic properties. It is shown that stiffness is an important performance factor. Previous models operate with links approximation and calculate stiffness matrix in the neighborhood of initial point. This is a reason why a new way for stiffness matrix calculation is proposed. This method is illustrated in a concrete industrial problem.

## 1 INTRODUCTION

Currently, parallel manipulators have become more and more popular for a variety of technological processes, including high-speed precision machining [1] [2]. This growing attention is inspired by their essential advantages over serial manipulators, which have already reached the dynamic performance limits. In contrast, parallel manipulators are claimed to offer better accuracy, lower mass/inertia properties, and higher structural rigidity (i.e. stiffness-to-mass ratio) [3].

These features are induced by their specific kinematic structure, which resists to the error accumulation in kinematic chains and allows convenient actuators location close to the manipulator base. This makes them attractive for innovative robotic systems, but practical utilization of the potential benefits requires development of efficient stiffness analysis techniques, which satisfy the computational speed and accuracy requirements of relevant design procedures.

Generally, the stiffness analysis evaluates the effect of the applied external torques and forces on the compliant displacements of the end-effector. Numerically, this property is defined through the “stiffness matrix”, which gives the relation between the translational/rotational displacement and the static forces/torques causing this transition [4]. Similar to other manipulator properties (kinematical, for instance), the stiffness essentially depends on the force/torque direction and on the manipulator configuration [5].

Several approaches exist for the computation of the stiffness matrix, such as the Finite Element Analysis (FEA), the matrix structural analysis (MSA), and the virtual joint method (VJM). Each of methods has their own assumptions and in general approximate/linearized the stiffness model with adequate accuracy for the specific problem. Let us consider the recent modification of the VJM proposed by the authors [6]. It allows extending it to the over-constrained manipulators and applying it at any workspace point, including the singular ones. The method is based on a multidimensional lumped-parameter model that replaces the link flexibility by localized 6-dof virtual springs that describe both the linear/rotational deflections and the coupling between them. The spring stiffness parameters are evaluated using FEA modelling to take into account real shape of the manipulator components. This gives almost the same accuracy as FEA but with essentially lower computational effort because it eliminates re-meshing through the workspace.

In contrast to previous works, the novelty of this paper appears in the computation of the stiffness matrix for the loaded mode in the neighborhood of static equilibrium with true dimensions links.

This paper focuses on the performance evaluation of the milling process for the composite materials via enhancing the precise stiffness model for the compliance errors estimation. VJM is used for the stiffness modeling. It incorporates accurate stiffness properties of links via improved FEA-based stiffness calculation algorithm. This leads to adequate estimation of deflections and allows improving accuracy of manufacturing via accurate planning of the technological process.

The reminder of the paper is organized as follows. In the Section 2, it is defined the set of problems which are considered in the paper. Section 3 presents performance evaluation factors. Section 4 proposes a new method for the stiffness modeling of parallel manipulator. Section 5 illustrate the efficiency of the propose technique on the Orthoglide manipulator by specializing it for milling of bathroom component. And, finally, section 6 summarizes the main contributions of the paper.

## 2 PROBLEM STATEMENT

The main problem to be solved in this article is the accuracy evaluation during the manufacturing step for specific parallel manipulators and technological process. Position accuracy and performances characteristics for the industrial manipulators depend on the technological process, kinetostatic and geometry properties of both manipulator and workpiece. Therefore specializing of the manipulator for the current manufacturing process with respect to the required performances properties allows increasing the position accuracy of the tool.

Mechanical stiffness is one of the most important properties, which have influence on the position accuracy of low mass parallel mechanisms. As it is mentioned in [7] Cartesian stiffness is a nonlinear function of the external loading and depends on the configuration of the manipulator and, as result, of the position of end effector in the workspace [8]. The computed Cartesian stiffness both depends on the stiffness of the links and stiffness model accuracy.

So the aim of this article is to specialize a parallel manipulator for the milling of products made of composite material with known geometric sizes. The idea is to improve the accuracy for the high-speed milling via mass reduction and improving of the stiffness model.

## 3 PERFORMANCE EVALUATION

Let us present here the most essential technology-oriented performance measures that are used in the design process of a machine-tool. They may be classified into two groups:

- *the measures based on geometric, kinematic and kinetostatic properties*; these measures are based on simple models that evaluate the geometric, kinematic and kinetostatic properties of a mechanism. Their specificity is that they only use the primary geometric parameters of the mechanism, i.e., the size of the base, the length of links, etc.
- *the measures based on dynamic, elastostatic and elastodynamic properties*; these measures mostly use more complicated models. Their specificity is that they not only use the primary geometric parameters of the mechanism, but also the secondary geometric parameters, as for example, the cross-section of links.

Let us now describe them.

### 3.1 Geometric, kinematic and kinetostatic performances

The most commonly used geometric, kinematic and kinetostatic performances are:

- *the velocity transmission factors*;

For appreciating the speed capability of a manipulator, several kinematic performance indices are defined using the Jacobian matrix **J** (see [10]), such as the condition number, the largest/smallest singular value (also called the maximal/minimal transmission factor and denoted as $k_v^{\max}$ and $k_v^{\min}$, respectively – Figure 1a), the dexterity, the manipulability, etc. However, as mentioned by Merlet in [10], the previously cited indices does not take into account the 'technological reality' of the mechanism, as they are based on the use of the Euclidian norm of the input velocity vector $\dot{\mathbf{\Phi}}$ ($\|\dot{\mathbf{\Phi}}\|$ being considered equal to 1) while it is clear that each actuator may have a velocity $\dot{\phi}_i \in [-\dot{\phi}_i^{\max}, \dot{\phi}_i^{\max}]$, where $\dot{\phi}_i$ and $\dot{\phi}_i^{\max}$ are the actual and maximal velocities for the actuator *i* (*i* = 1 to *n*). Thus, it is necessary to redefine the transmission factors.

Two types of transmission factors may be used: (i) the velocity transmission factors along all directions of the workspace; in such a case, the unit square is mapped into a parallelepiped (Figure 1b) and (ii) the velocity transmission along some particular directions of the workspace; in such a case, the unit square is mapped into a hexagon (Figure 1c) [9].

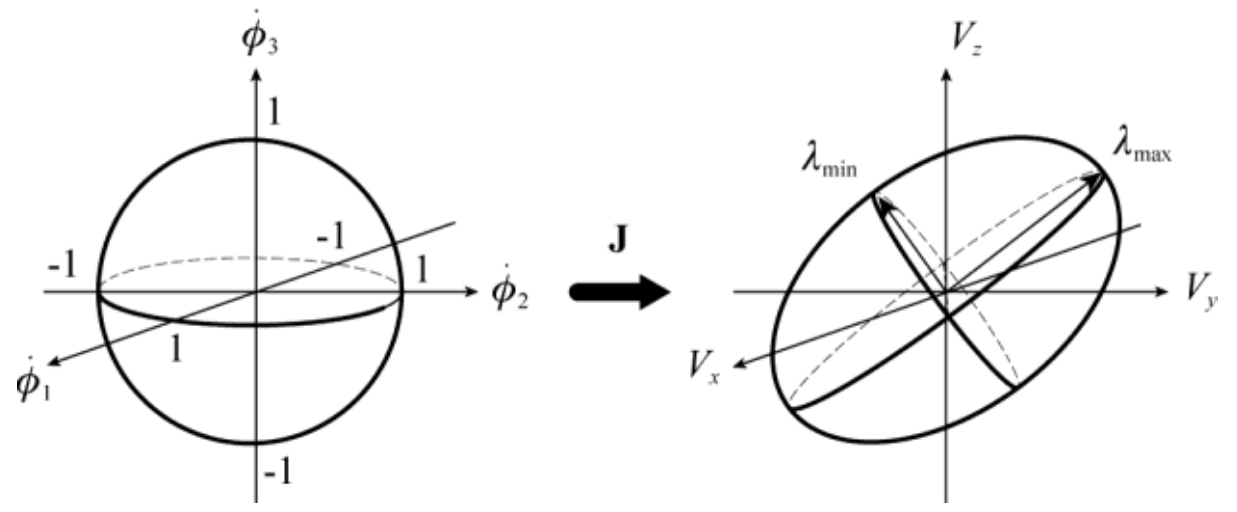

(a) mapping of the unit sphere

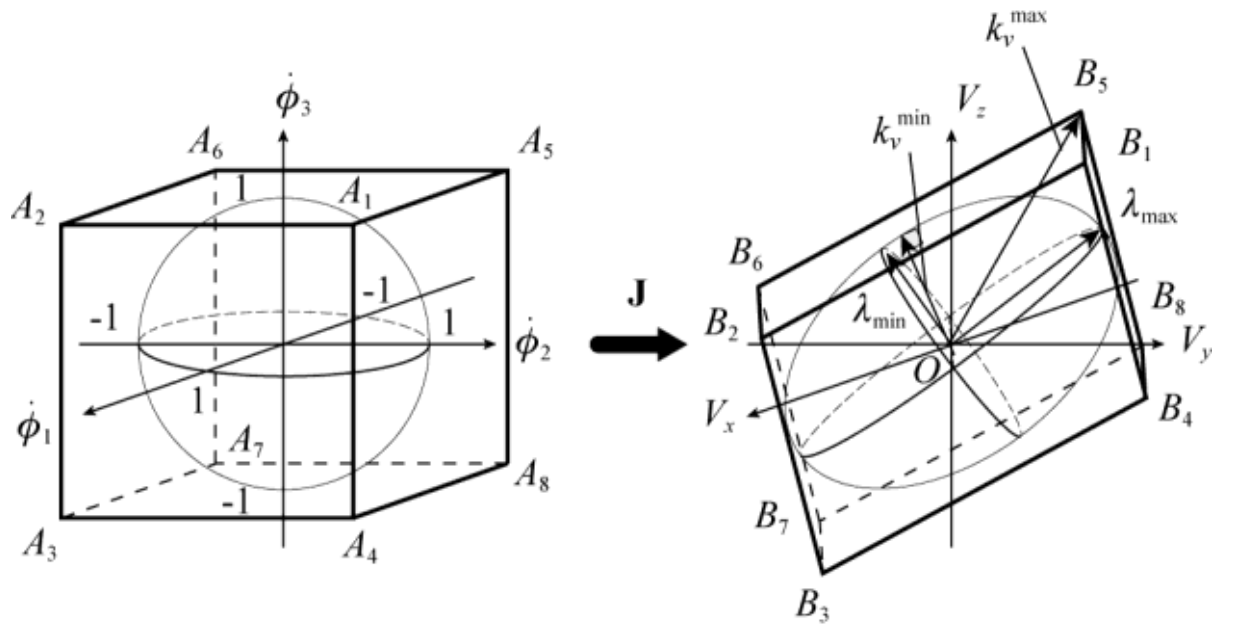

(b) mapping of the unit cube.

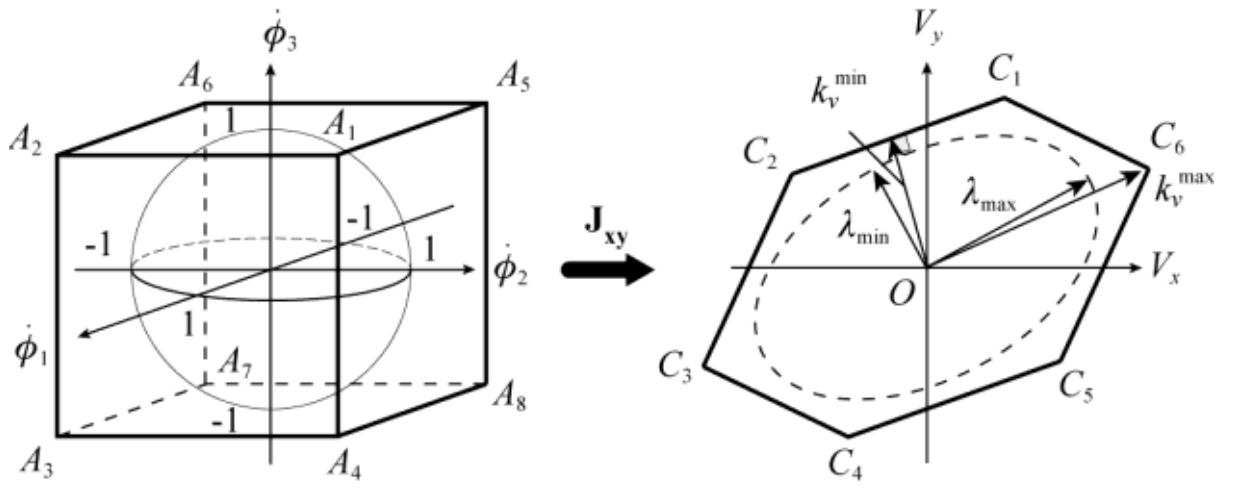

(c) mapping of the velocity transmission in the xy plane.

Figure 1: Mapping, using the Jacobian matrix.

- *the accuracy transmission factors*;

The accuracy of a mechanism may be due to different factors. However, as pointed out by Merlet [11], active-joint errors are the most significant source of errors in a properly designed, manufactured, and calibrated parallel robot. The classical approach consists in considering the first order approximation that maps the input error to the output error:

$$\delta \mathbf{p} = \mathbf{J}\, \delta \mathbf{\Phi} \quad (1)$$

where $\delta\mathbf{\Phi}$ represents the vector of the active-joint errors, $\delta\mathbf{p}$ the vector of end-effector errors. This method will give only an approximation of the end-effector maximum error, but at an optimisation stage, this model may be sufficient.

It is possible to show that, due the use of the expression (1) as the accuracy model, the accuracy transmission factors are similar to the velocity transmission factors. More information about the computation of these transmission factors is proposed in [9].

- *the force transmission factors*;

For 3-DOF translational parallel manipulators, the input efforts $\boldsymbol{\tau}$ are related to the forces **f** applied on the platform by the following relation:

$$\mathbf{f} = \mathbf{J}^{-T}\,\boldsymbol{\tau} \qquad (2)$$

Looking at expression (2), it appears that the relationship for the accuracy is similar to the relationship for the velocity, but $\mathbf{J}^{-T}$ is used instead to **J**. Moreover, it is also clear that the effort $\tau_i$ of one actuator is comprised between $-\tau^{\max}$ and $+\tau^{\max}$, $\tau^{\max}$ being the maximal effort admissible by the actuated pair. So, the force transmission factors may be computed in the same manner as the velocity transmission factors. The maximal and minimal force and moment transmission factors will be denoted as $k_f^{\min}$ and $k_f^{\max}$ respectively. To have more information about the computation of these transmission factors, the reader is proposed to refer to [9].

- *the size of the workspace*;

Using the set of geometric parameters, the workspace **W** may be generated using the kinematic equations and the (passive and active) joint limits [3]. Since, for the considered application, the desired regular workspace is a parallelepiped $\mathbf{W_0}$ of size $\{a_0 \times b_0 \times c_0\}$, the relevant measure may be defined by the largest similar object $\mathbf{W}^{abc} = \{\mu a_0 \times \mu b_0 \times \mu c_0\}$ inscribed in **W**, i.e.

$$\mathbf{W}^{abc} = \mathbf{T}(\mu\,\mathbf{W_0}); \quad (\mu,\mathbf{T}) = \arg\max_{\mu,\mathbf{T}} \left\{\mu \,\middle|\, \mathbf{T}(\mu\,\mathbf{W_0}) \subset \mathbf{W}\right\} \qquad (3)$$

where $\mu$, **T** are respectively the scalar scaling factor and the coordinate transformation operator in the Cartesian space. An algorithm that is able to compute the size largest parallelepiped inside a given workspace is presented in [9]

Let us now introduce the dynamic, elastostatic and elastodynamic performances.

### 3.2 Dynamic, elastostatic and elastodynamic performance

The most commonly used dynamic, elastostatic and elastodynamic performances are:

- *the total mass of the robot*;

This is probably the simplest performance measure. The masses of the different elements of a robot have a direct influence on its dynamic behaviour. For one given structure with fixed value of links length, the robot that will have the smallest mass will be the one with the smallest input efforts, and as a result, the highest acceleration capacity.

- *the input efforts*;

The input efforts, denoted as $\boldsymbol{\tau}$, depend on the mass and axial moment of inertia of links, such as friction in joints and position, velocity and acceleration of the robot. There expression is given by:

$$\boldsymbol{\tau} = \mathbf{M}(\mathbf{q})\ddot{\mathbf{q}} + \mathbf{C}(\mathbf{q},\dot{\mathbf{q}})\dot{\mathbf{q}} + \mathbf{G}(\mathbf{q}) + \mathbf{J}^T\mathbf{f} \qquad (4)$$

where **M** is the mass matrix, **C** the matrix of the Coriolis, centrifugal and viscous friction effects, **G** is the matrix of the gravity and Coulomb friction effects. **q**, $\dot{\mathbf{q}}$ and $\ddot{\mathbf{q}}$ represent the vectors of the positions, velocities and accelerations of the actuators. **f** is an external force applied on the platform. Generally, it is considered that, given a desired trajectory, the robot that has the lowest input efforts along the trajectory has the best performance.

- the maximal deformations;

For parallel manipulators, elasticity is an essential performance measure since it is directly related to the positioning accuracy and the payload capability. Mathematically, this benchmark is defined by the stiffness matrix **K**, which describes the relation between the linear/angular displacements $\delta\mathbf{t}$ of the end-effector and the external forces/torques **f** applied on the tool:

$$\mathbf{f} = \mathbf{K}\,\delta\mathbf{t} \qquad (5)$$

It is obvious that elasticity is highly dependent upon geometry, materials and link shapes that are completely defined with the CAD model. The stiffness matrix may be computed using three methods: the finite element analysis (FEA) [12, 13], the matrix structural analysis (SMA) [14, 15] and the virtual joint method (VJM) [6, 16, 17].

Whatever will be the way to compute the deformations of the robot, the mechanism that will have the best elastostatic performances will be the one that will have the smallest deformations under a given force, along a specific trajectory or in the totality (or some portions) of the workspace.

External forces are caused by the coupling of the tool and machining piece. But previous stiffness models calculate stiffness matrix in the neighborhood of initial point and operate with links approximations. Such models can't guarantee good accuracy. Therefore let us consider the stiffness model improvement and deflections evaluation.

## 4 STIFFNESS MODEL

### 4.1 Problem of stiffness modeling

To evaluate the manipulator stiffness, let us apply the VJM method that assumes that the traditional rigid model is extended by adding virtual joints, which describe stiffness of the actuator and links. Thus, the end-effector position for each chain of the manipulator can be described as

$$\mathbf{t} = \mathbf{g}(\mathbf{q},\boldsymbol{\theta}) \qquad (6)$$

where $\mathbf{g}(...)$ is the geometry function which depends on the passive $\mathbf{q}$ and virtual joint $\boldsymbol{\theta}$ coordinates, the vectors $\mathbf{q} = (q_1, q_2, ..., q_n)^T$ includes all passive joint coordinates, the vector $\boldsymbol{\theta} = (\theta_1, \theta_2, ..., \theta_m)^T$ collects all virtual joint coordinates, $n$ is the number of passive joins, $m$ is the number of virtual joints.

This expression includes both traditional geometric variables (passive and active joint coordinates) and stiffness variables (virtual joint coordinates).

To evaluate the manipulator ability to respond to external forces and torques, it is necessary to introduce additional equations that define the virtual joint reactions to the corresponding spring deformations. For analytical convenience, corresponding expressions may be collected in a single matrix equation

$$\boldsymbol{\tau}_{\boldsymbol{\theta}} = \mathbf{K}_{\boldsymbol{\theta}} \cdot \boldsymbol{\theta} \qquad (7)$$

where $\boldsymbol{\tau}_{\boldsymbol{\theta}} = \left(\tau_{\theta,1}, \tau_{\theta,2}, ..., \tau_{\theta,m}\right)^T$ is the aggregated vector of the virtual joint reactions, $\mathbf{K}_{\boldsymbol{\theta}} = diag\left(\mathbf{K}_{\boldsymbol{\theta},1}, \mathbf{K}_{\boldsymbol{\theta},2}, ..., \mathbf{K}_{\boldsymbol{\theta},m}\right)$ is the aggregated spring stiffness matrix of the size $m\times m$, and $\mathbf{K}_{\boldsymbol{\theta},i}$ is the spring stiffness matrix of the corresponding link.

For the compliant link, the matrix **K** can be computed using the FEA-based techniques, which usually produce rather accurate result. Using the FEA, the stiffness matrix **K** (or its inverse **k**) is evaluated from several numerical experiments, each of which produces the vectors of linear and angular deflections (**t**, $\boldsymbol{\varphi}$) corresponding to the applied force and torque (**F**, **M**). Then, the desired matrix is computed from the linear system

$$\mathbf{k} = \begin{bmatrix} \mathbf{F}_1 & \dots & \mathbf{F}_m \\ \mathbf{M}_1 & \dots & \mathbf{M}_m \end{bmatrix}^+ \cdot \begin{bmatrix} \mathbf{t}_1 & \dots & \mathbf{t}_m \\ \boldsymbol{\varphi}_1 & \dots & \boldsymbol{\varphi}_m \end{bmatrix} \quad (8)$$

where $[\ ]^+$ is pseudoinverse of the rectangle matrix, *m* is the number of experiments ($m \ge 6$) and the matrix pseudoinverse is replaced by the inverse in the case of $m = 6$. It is obvious that this case with special arrangement of the forces and torques is numerically attractive (for more detail see [18]).

In order to increase accuracy it is worth to improve the deflection estimation technique. It is proposed to evaluate (**t**, $\boldsymbol{\varphi}$) from the *displacement field* describing transitions of rather large number of nodes located in the neighborhood of reference point (RP).

To formulate this problem strictly, let us denote the displacement field by a set of vector couples $\{\mathbf{p}_i, \Delta\mathbf{p}_i \mid i = 1,2,...,n\}$ where the first component $\mathbf{p}_i$ define the node initial location (before applying the force/torque), $\Delta\mathbf{p}_i$ refers to the node displacement due to the applied force/torque, and *n* is the number of considered nodes. Then, assuming that all the nodes are close enough to the reference point, this set can be approximated by a "rigid transformation"

$$\mathbf{p}_i + \Delta\mathbf{p}_i = \mathbf{R}(\boldsymbol{\varphi}) \cdot \mathbf{p}_i + \mathbf{t}\,, \quad i = 1,2,...,n \quad (9)$$

that includes as the parameters the linear displacement **t** and the orthogonal 3×3 matrix **R** that depends on the rotational displacement $\boldsymbol{\varphi}$. Then, the problem of the deflection estimation can be presented as the best fit of the considered vector field by equation (4) with respect to six scalar variables incorporated in **t**, **R**.

In general case, the desired stiffness model is defined by a non-liner relation

$$\mathbf{F} = \xi(\Delta\mathbf{t}) \quad (10)$$

that describes resistance of a mechanism to deformations $\Delta\mathbf{t}$ caused by an external force/torque $\mathbf{F}$ [1]. It should be noted that the mapping $\Delta\mathbf{t} \to \mathbf{F}$ is strictly mathematically defined and physically tractable in all cases, including under-constrained kinematics and singular configurations of the manipulator. However, the converse is not true.

In engineering practice, function $\xi(...)$ is usually linearized in the neighborhood of the static equilibrium $(\mathbf{q}, \boldsymbol{\theta})$ corresponding to the end-effector position $\mathbf{t}$ and external loading $\mathbf{F}$. For the unloaded mode, i.e. when $\mathbf{F} = \mathbf{0}$ and $\boldsymbol{\theta}_0 = \mathbf{0}$ the stiffness model is expressed by a simple relation

$$\mathbf{F} \approx \mathbf{K}(\mathbf{q}_0, \boldsymbol{\theta}_0) \cdot \Delta\mathbf{t} \quad (11)$$

where **K** is 6×6 "stiffness matrix" and the vector $\mathbf{q}_0 = (q_{01}, q_{02}, ..., q_{0n})^T$ defines the equilibrium configuration corresponding to the end-effector location $\mathbf{t}_0$, in accordance with the manipulator geometry.

However, for the loaded mode, stiffness model have to be defined in the neighborhood of the static equilibrium that corresponds to another manipulator configuration $(\mathbf{q}, \boldsymbol{\theta})$, which is caused by external forces $\mathbf{F}$. In this case, the stiffness model describes the relation between the increments of the force $\delta\mathbf{F}$ and the position $\delta\mathbf{t}$

$$\delta\mathbf{F} \approx \mathbf{K}(\mathbf{q}, \boldsymbol{\theta}) \cdot \delta\mathbf{t} \quad (12)$$

where $\mathbf{q} = \mathbf{q}_0 + \Delta\mathbf{q}$ and $\boldsymbol{\theta} = \boldsymbol{\theta}_0 + \Delta\boldsymbol{\theta}$ denote the new position of the manipulator, $\Delta\mathbf{q}$ and $\Delta\boldsymbol{\theta}$ are the deviations of the passive joint and virtual spring coordinates.

Hence, the problem of the stiffness modeling in the loaded mode may be divided into two sequential subtasks: (i) stiffness model identification from the vector field of displacements and (ii) linearization of relevant force/position relations in the neighborhood of the loaded configuration. Let us consider these two sub-problems consequently.

### 4.2 Stiffness model identification

To estimate the desired deflections (**t**, $\boldsymbol{\varphi}$), let us apply the least square technique that leads to minimization of the sum of squared residuals

$$f = \sum_{i=1}^{n} \left\| \mathbf{p}_i + \Delta\mathbf{p}_i - \mathbf{R}(\boldsymbol{\varphi})\mathbf{p}_i - \mathbf{t} \right\|^2 \to \min_{\mathbf{R},\mathbf{t}} \quad (13)$$

with respect to the vector **t** and the orthogonal matrix **R** representing the rotational deflections $\boldsymbol{\varphi}$**.** The specificity of this problem (that does not allow direct application of the standard methods) are the orthogonally constraint $\mathbf{R}^T\mathbf{R} = \mathbf{I}$ and non-trivial relation between elements of the matrix **R** and the vector $\boldsymbol{\varphi}$**.** To reduce the computational efforts, let us linearize the rotational matrix **R** [18]. This allows to rewrite equation of the 'rigid transformation' (9) in the form

$$\Delta\mathbf{p}_i = \mathbf{p}_i \times \boldsymbol{\varphi} + \mathbf{t}; \quad i = \overline{1,n} \quad (14)$$

that can be further transformed into a linear system of the following form

$$\begin{bmatrix} \mathbf{I} & \mathbf{P}_i \end{bmatrix} \begin{bmatrix} \mathbf{t} \\ \boldsymbol{\varphi} \end{bmatrix} = \Delta\mathbf{p}_i; \quad i = \overline{1,n} \quad (15)$$

where $\mathbf{P}_i$ is a skew-symmetric matrix corresponding to the vector $\mathbf{p}_i$. Then, applying the standard least-square technique and shifting the origin of the coordinate system to the point $\mathbf{p}_c = n^{-1}\sum_{i=1}^{n}\mathbf{p}_i$ leading to expression

$$\begin{bmatrix} \mathbf{t} \\ \boldsymbol{\varphi} \end{bmatrix} = \begin{bmatrix} n^{-1}\mathbf{I} & \mathbf{0} \\ \mathbf{0} & \left( \sum_{i=1}^{n} \hat{\mathbf{P}}_i^T \hat{\mathbf{P}}_i \right)^{-1} \end{bmatrix} \cdot \begin{bmatrix} \sum_{i=1}^{n} \Delta\mathbf{p}_i \\ \sum_{i=1}^{n} \hat{\mathbf{P}}_i^T \Delta\mathbf{p}_i \end{bmatrix} \quad (16)$$

that requires inversion of the matrix of size 3×3. Here, following the adopted notation $\hat{\mathbf{P}}_i$ is a skew-symmetric matrix corresponding to the vector $\hat{\mathbf{p}}_i = \mathbf{p}_i - \mathbf{p}_c$.

By its general principle, the FEA-modeling is an approximate method that produces some errors caused by the discretization. Beside, even for the perfect modeling, the deflections in the neighborhood of the reference point do not exactly obey the equation (4). Hence, it is reasonable to assume that the 'rigid transformation' (9) incorporates some random errors

$$\mathbf{p}_i + \Delta\mathbf{p}_i = \mathbf{R}(\varphi) \cdot \mathbf{p}_i + \mathbf{t} + \boldsymbol{\varepsilon}_i; \quad i = \overline{1,n} \quad (17)$$

that are supposed to be independent and identically distributed Gaussian random variables with zero-mean and standard deviation $\sigma$.

In the frame of this assumption, the expression for the deflections (16) can be rewritten as

$$\mathbf{t} = \mathbf{t}^{o} + n^{-1}\sum_{i=1}^{n}\boldsymbol{\varepsilon}_i; \quad \boldsymbol{\varphi} = \boldsymbol{\varphi}^{o} + \left(\sum_{i=1}^{n}\hat{\mathbf{P}}_i^T\hat{\mathbf{P}}_i\right)^{-1}\sum_{i=1}^{n}\hat{\mathbf{P}}_i^T\boldsymbol{\varepsilon}_i \qquad (18)$$

where the superscript 'o' corresponds to the 'true' parameter value. This justifies usual properties of the adopted point-type estimator (16), which is obviously unbiased and consistent. Furthermore, the variance-covariance matrices for **t**, φ may be expressed as

$$\mathrm{cov}[\mathbf{t}] = \frac{\sigma^2}{n}\mathbf{I}; \quad \mathrm{cov}[\boldsymbol{\varphi}] = \sigma^2\left(\sum_{i=1}^{n}\hat{\mathbf{P}}_i^T\hat{\mathbf{P}}_i\right)^{-1} \qquad (19)$$

allowing to evaluate the estimation accuracy using common confidence interval technique.

Another practical question is related to *detecting zero elements* in the compliance matrix or, in other word, evaluating the statistical significance of the computed values compared to zero. Relevant statistical technique [19] operates with the p-values that may be easily converted in the form $k\,\sigma_a$ , where k is usually from 3 to 5 and the subscript '$a$' refers to a particular component of the vectors **t**, **φ.**

To evaluate the standard deviation σ describing the random errors $\boldsymbol{\varepsilon}$, one may use the residual-based estimator obtained from the expression

$$\mathrm{E}\left(\sum_{i=1}^{n}\left\|\mathbf{p}_i + \Delta\mathbf{p}_i - \mathbf{R}(\boldsymbol{\varphi})\cdot\mathbf{p}_i - \mathbf{t}\right\|^2\right) = (3n-6)\,\sigma^2 . \qquad (20)$$

The latter may be easily derived taking into account that, for each experiment, the deflection filed consist of $n$ three-dimensional vectors that are approximated by the model containing 6 scalar parameters. Moreover, to increase accuracy, it is prudent to aggregate the squared residuals for all FEA-experiments and to make relevant estimation using the coefficient $(3n-6)\,m\,\sigma^2$ , where $m$ is the experiments number.

In addition, to increase accuracy and robustness, it is reasonable to eliminate outliers in the experimental data. They may appear in the FEA-field due to some anomalous causes, such as insufficient meshing of some elements, violation of the boundary conditions in some areas of the mechanical joints, etc. The simplest and reliable method that is adopted in this paper is based on the 'data filtering' with respect to the residuals.

### 4.3 Stiffness model in the loaded mode

To compute the desired stiffness matrix, let us consider the neighborhood of the loaded configuration and assume that the external force and the end-effector location are incremented by some small values $\delta\mathbf{F}$ , $\delta\mathbf{t}$ . Besides, let us assume that a new configuration satisfies the equilibrium conditions [7]. Hence, it is necessary to consider simultaneously two equilibriums corresponding to the manipulator state variables $(\mathbf{F},\mathbf{q},\boldsymbol{\theta},\mathbf{t})$ and $(\mathbf{F}+\delta\mathbf{F},\mathbf{q}+\delta\mathbf{q},\boldsymbol{\theta}+\delta\boldsymbol{\theta},\mathbf{t}+\delta\mathbf{t})$ . Relevant equations of statics may be written as

$$\mathbf{F}\cdot\mathbf{J}_\theta^{\,T} = \mathbf{K}_\theta\cdot\boldsymbol{\theta}; \qquad \mathbf{F}\cdot\mathbf{J}_q^{\,T} = 0 \qquad (21)$$

and

$$\begin{aligned}&(\mathbf{F}+\delta\mathbf{F})\cdot(\mathbf{J}_\theta+\delta\mathbf{J}_\theta)^T = \mathbf{K}_\theta\cdot(\boldsymbol{\theta}+\delta\boldsymbol{\theta});\\&(\mathbf{F}+\delta\mathbf{F})\cdot(\mathbf{J}_q+\delta\mathbf{J}_q)^T = 0\end{aligned} \qquad (22)$$

where $\delta\mathbf{J}_q(\mathbf{q},\boldsymbol{\theta})$ and $\delta\mathbf{J}_\theta(\mathbf{q},\boldsymbol{\theta})$ are the differentials of the Jacobians due to changes in $(\mathbf{q},\boldsymbol{\theta})$. Besides, in the neighborhood of $(\mathbf{q},\boldsymbol{\theta})$ , the kinematic equation may be also presented in the linearized form:

$$\boldsymbol{\delta}\mathbf{t} = \mathbf{J}_\theta(\mathbf{q},\boldsymbol{\theta})\cdot\boldsymbol{\delta\theta} + \mathbf{J}_q(\mathbf{q},\boldsymbol{\theta})\cdot\boldsymbol{\delta}\mathbf{q}\ , \qquad (23)$$

Hence, after neglecting the high-order small terms and expending the differentials via the Hessians of the function $\Psi = \mathbf{g}(\mathbf{q},\boldsymbol{\theta})^T\mathbf{F}$ , equations (21), (22) may be rewritten as

$$\begin{aligned}&\mathbf{J}_\theta^{\,T}(\mathbf{q},\boldsymbol{\theta})\cdot\delta\mathbf{F} + \mathbf{H}_{\theta q}^F(\mathbf{q},\boldsymbol{\theta})\cdot\delta\mathbf{q} + \mathbf{H}_{\theta\theta}^F(\mathbf{q},\boldsymbol{\theta})\cdot\delta\boldsymbol{\theta} = \mathbf{K}_\theta\cdot\delta\boldsymbol{\theta}\\&\mathbf{J}_q^{\,T}(\mathbf{q},\boldsymbol{\theta})\cdot\delta\mathbf{F} + \mathbf{H}_{qq}^F(\mathbf{q},\boldsymbol{\theta})\cdot\delta\mathbf{q} + \mathbf{H}_{q\theta}^F(\mathbf{q},\boldsymbol{\theta})\cdot\delta\boldsymbol{\theta} = \mathbf{0}\end{aligned} \qquad (24)$$

and the general relation for the stiffness matrix in the loaded mode can be presented as

$$\begin{bmatrix}\mathbf{J}_\theta\mathbf{k}_\theta^F\mathbf{J}_\theta^T & \mathbf{J}_q + \mathbf{J}_\theta\mathbf{k}_\theta^F\mathbf{H}_{\theta q}^F\\ \mathbf{J}_q^T + \mathbf{H}_{q\theta}^F\mathbf{k}_\theta^F\mathbf{J}_\theta^T & \mathbf{H}_{qq}^F + \mathbf{H}_{q\theta}^F\mathbf{k}_\theta^F\mathbf{H}_{\theta q}^F\end{bmatrix}^{-1} = \begin{bmatrix}\mathbf{K}_F & \mathbf{K}_q\\ * & *\end{bmatrix} \qquad (25)$$

Hence, the presented technique allows computing the stiffness matrix in the presence of the external load and to generalize previous results both for serial kinematic chains and for parallel manipulators.

## 5 APPLICATION EXAMPLE

### 5.1 Industrial problem

Let us demonstrate the efficiency of our design approach on a concrete problem coming from the industrial sector of the region of Nantes (France). One of the most important activity areas of this region is the manufacturing of bathroom components (shower cabin, washbasin, bathtub, etc.). Most of parts used during the assembly process are made of thermosetting materials. The main operations achieved on these parts is trimming, i.e. the suppression of the edges of the parts in order to obtain a good surface roughness.

The tools used for milling are specific mills, which are composed of a large number a diamond glued on its surface. Therefore the number of tooth is supposed to be infinite. However, the use of these specific tools allows the simplification of the model for computing the milling forces.

Let us consider the trimming operation depicted at Figure 2. The milling force may be decomposed into two components, denoted as $F_t$ (the tangential force) and $F_f$ (the radial force) (the vertical component can be neglected for such kind of operation

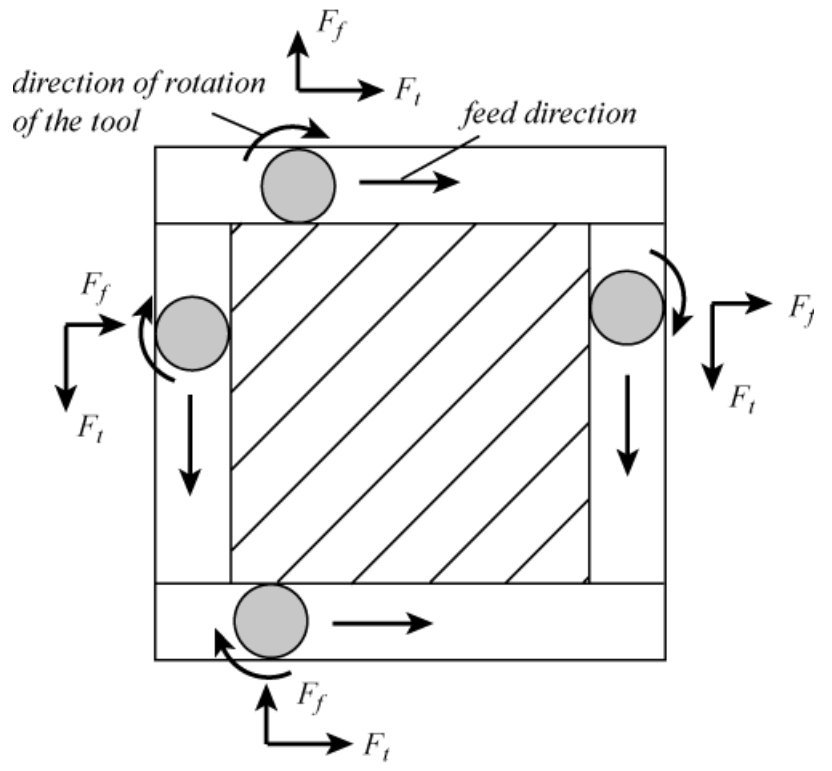

Figure 2: Trimming operation for composite material.

The machines tools that are used for the trimming of the bathroom components proposed on the Figure 3 must be designed such as they attain the following characteristics:

- workspace $\mathbf{W}^{abc}$ of size {0.5 m × 0.5 m × 0.5 m};
- $\|\mathbf{v}_{xy}\|$ = 60 m/min ($\mathbf{v}_{xy}$ contains the components of the platform velocity vector **v** in the *xy* plane);
- $\|\mathbf{f}_{xy}\|$ = 300 N ($\mathbf{f}_{xy}$ contains the components of the external effort vector **f** in the *xy* plane);
- $\|\delta\mathbf{p}_{xy}\|$ = 0.25 mm ($\delta\mathbf{p}_{xy}$ contains the components of the platform deformations vector $\delta\mathbf{p}$ in the *xy* plane).

### 5.2 Architecture of the manipulator

For the milling process let us specialize the Orthoglide manipulator (Figure 4) [20]. This architecture was built in Institut de Recherche en Communications et Cybernetique de Nantes (IRCCyN) and satisfies the following design objectives: cubic Cartesian workspace of size 200×200×200 mm$^3$ (while for selection treatment required workspace 200×200×200 mm$^3$), Cartesian velocity and acceleration in the isotropic point of 1.2 m/s and 14 m/s$^2$; payload of 4 kg; transmission factor range 0.5–2.0. The legs nominal geometry is defined by the following parameters: $L$ = 310 mm, $d$ = 100 mm, $r$ = 31 mm where $L$, $d$ are the parallelogram length and width, and $r$ is the distance between the kinematic parallelogram and the tool centre point. Stiffness model (elements and whole model) of Orthoglide manipulator is presented in [18].

### *5.3* Performance evaluation

Since the workspace of the Orthoglide manipulator is lower than required for milling, the length and cross-section of the bar element was increased. Using proposed technique for the stiffness model identification, the new stiffness matrix for the bar element was obtained (Table 1).

Now let us compute the displacements caused by the forces during milling of the composite material. The three subtasks are considered: estimation of the force size, estimation of the displacement through the whole workspace and analyzing of the force direction.

***Force direction analysis***. Here let us plot errors for the constant force (300 N) rotating it from -180 to 180 deg for four typical work points: original configuration (x=y=z=0), two opposite corners of workspace x=y=z=-200 and x=y=z=300 and point for nonsymmetrical configuration of the manipulator (x=-200, y=300, z=0). Results are presented on Figure 5. Figures 5a-d show the size of deflections for each direction of the force and represent it in polar coordinates, while Figures 5e-h show ellipses of deflections in the Cartesian coordinates. As we can see the compliance of the manipulator depends on the direction of external force and on the workpoint. Only for the isotropic work point (x=y=z=0) it is constant for each direction. The lowest and the highest compliant directions of the manipulator differ on $\pi/2$. For the work point x=y=z=-200, the manipulator is 7 times stronger in the direction 135/-45 deg than in the direction 45/-135 deg. The maximum compliance is 20% higher than for the single force (Fx, Fy) along the principal directions x and y of the frame. For the work point x=y=z=300, a force oriented along the principal axes of the frame causes deformations 20% less than for the worst direction and 50% more than for the strongest direction. It should be noted that in the workpoint (x=-200, y=300, z=0), forces along x and y directions cause different deformations. Along the x direction, the deformations are close to the minimum, while along the y direction, they are close to the maximum. The rate between the maximal and minimal deformations is about 7.

***Force size analyses***. Results for the same four workpoints are presented on Figure 6. The force-deflection relationship for the force less than 1000 N is linear, but depends on the manipulator configuration. Another conclusion is that in the work point x=y=z=0, the required accuracy can be satisfied for the force up to 600 N, while for all other tested workpoints, it can be satisfied only for forces inferior to 100 N. Moreover, taking into account the force direction, maximum compliance errors for the force 300 N may raise up to 1 mm and more.

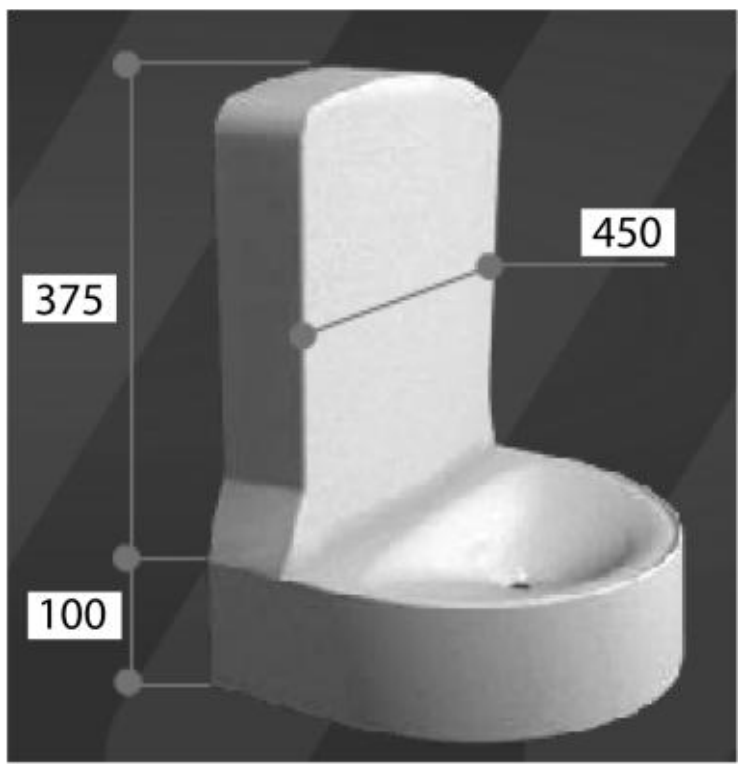

Figure 3: Typical examples of bathroom components manufactured in the region of Nantes (France).

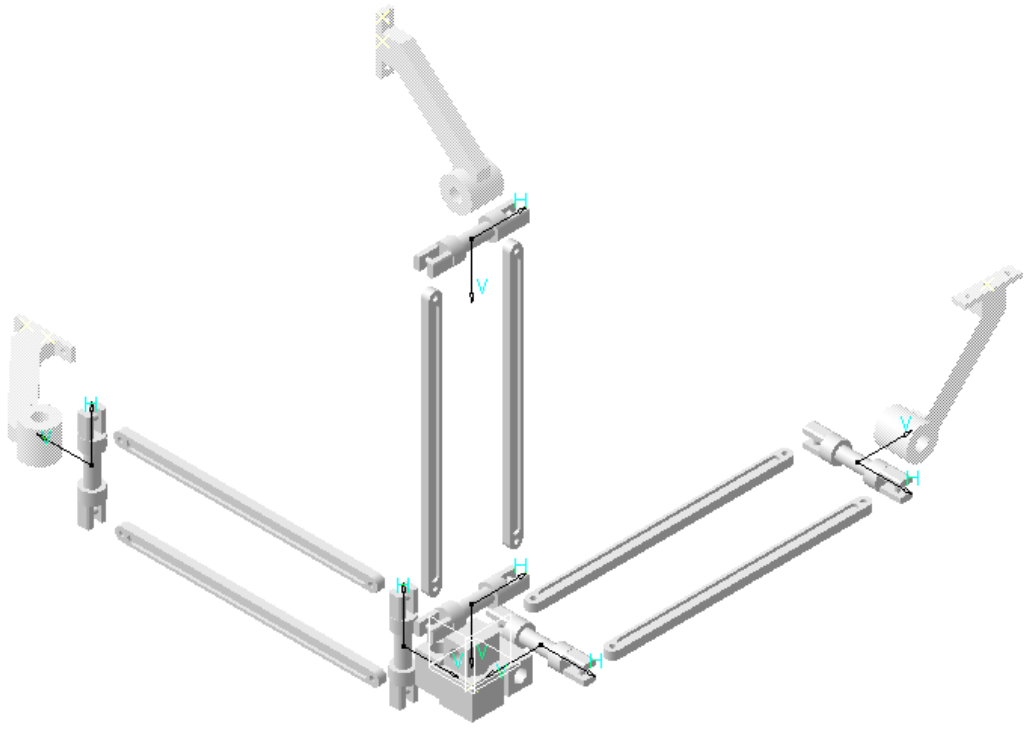

Figure 4: CAD model of Orthoglide manipulator

| Model | Compliance matrix elements | | | | | |
|---|---|---|---|---|---|---|
| | $k_{11}$ mm/N | $k_{22}$ mm/N | $k_{33}$ mm/N | $k_{44}$ rad/N·mm | $k_{55}$ rad/N·mm | $k_{66}$ rad/N·mm |
| Original Bar [18] | 4.55×10$^{-5}$ | 2.33×10$^{-1}$ | 5.08×10$^{-2}$ | 2.88×10$^{-5}$ | 1.50×10$^{-6}$ | 7.19×10$^{-6}$ |
| Revised Bar | 3.10×10$^{-5}$ | 3.54×10$^{-1}$ | 6.91×10$^{-2}$ | 0.39×10$^{-5}$ | 0.33×10$^{-6}$ | 1.74×10$^{-6}$ |

Table 1: Stiffness of bar element

**Workspace analysis.** For the accuracy control through the whole workspace, error maps for opposite planes (z = -200 mm Figure 7a and z = 300 mm Figure 7c) and for the “zero plane” (z = 0 mm Figure 7b) of workspace are presented. It should be noted that the position accuracy depends on the configuration of the manipulator and vary from 0,1 mm to more than 0.5 mm for z=-200 and z=300. The accuracy of the “zero plane” is satisfied for the operation, while for z=-200 mm, guaranteed accuracy is only 0.4 mm, and for z=300, 0.5 mm. But milling in the strongest direction of the manipulator leads to an accuracy more than 0,2 mm. So, optimizing the milling process for the specialized Orthoglide manipulator may improve the accuracy more than 3 times.

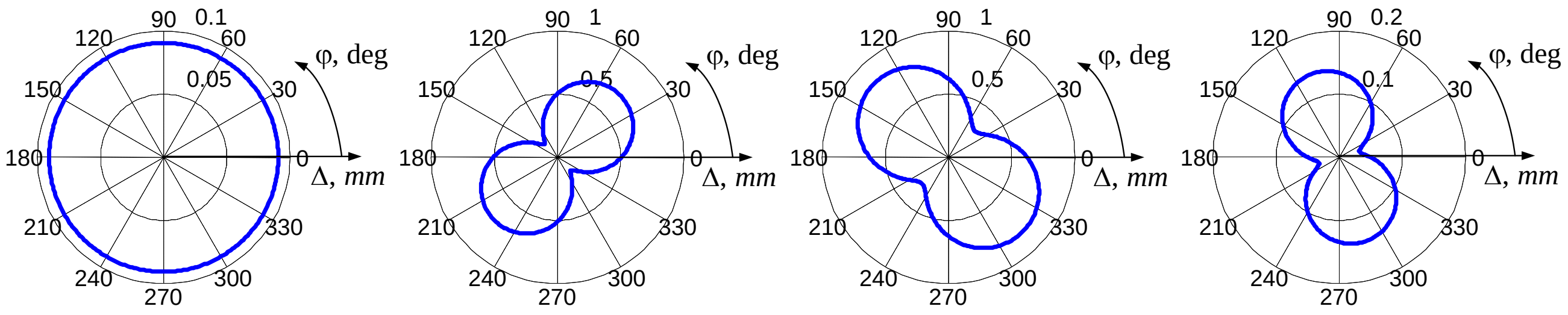

(a) x=0, y=0, z=0 (b) x=-200, y=-200, z=-200 (c) x=300, y=300, z=300 (d) x=-200, y=300, z=0

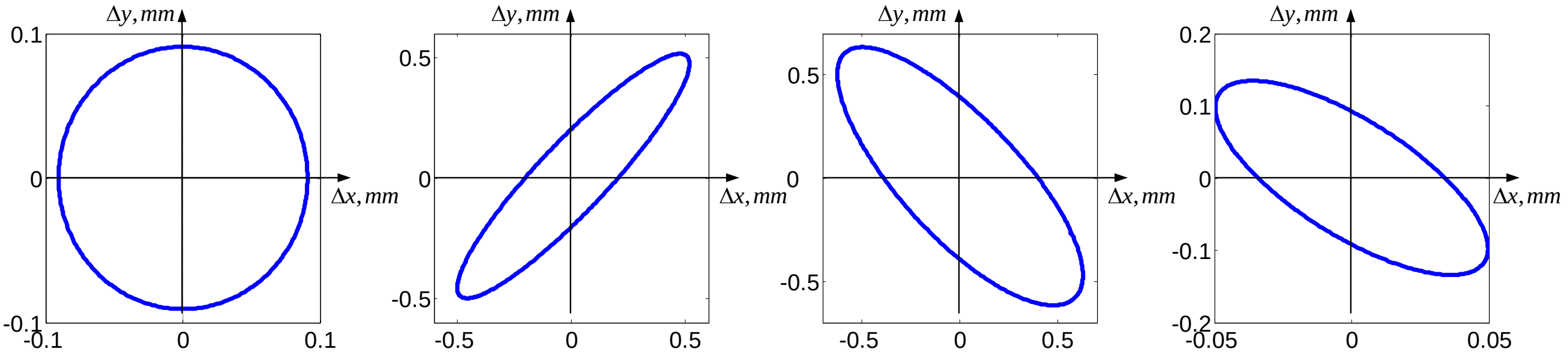

(e) x=0, y=0, z=0 (f) x=-200, y=-200, z=-200 (g) x=300, y=300, z=300 (h) x=-200, y=300, z=0

Figure 5: Norm of the deformations of the end-effector as a function of the direction φ phi of the planar force F= 300 N: (a)-(d) in the polar coordinates ( $\Delta = \sqrt{\Delta x^2 + \Delta y^2}$ ); (e)-(h) in the Cartesian coordinates

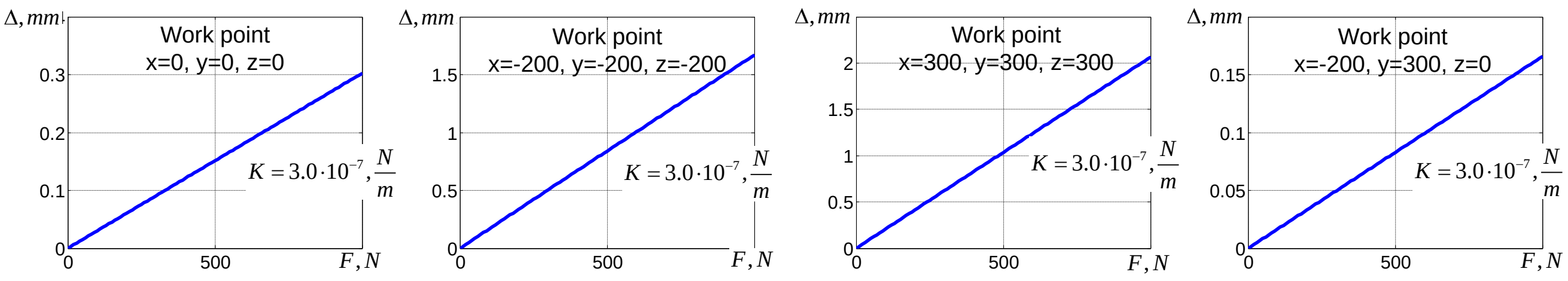

(a) (b) (c) (d)

Figure 6: Force-Deflections relationship in the four test points

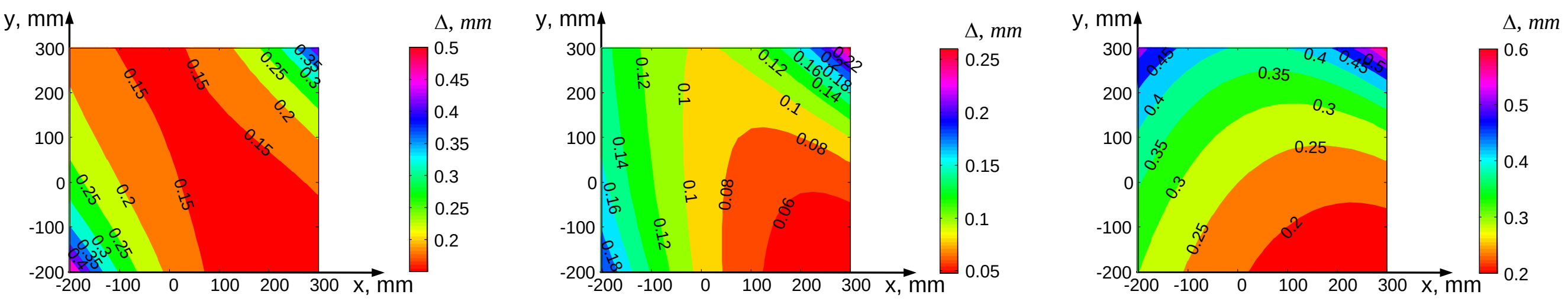

(a) (b) (c)

Figure 7: Error maps for the revised manipulator

## 6 SUMMARY

The accuracy of milling of composite materials depends on the number of factors such as accurate kinematic and stiffness modeling, performance evaluation, force control, planning of milling process and others. This paper contributes to the methodology, which is used for the accurate stiffness modeling for the manipulator and it links for the estimation of the deflections errors. It allows evaluating compliance errors caused by the technological process and detecting strongest and lowest directions for the compliance errors. Using this information for the planning of the milling process allows increasing the accuracy of processing of the pieces made of composite materials.

The method is efficiently illustrated on the milling of bathroom components produced in the Nantes region. For the manufacturing, Orthoglide manipulator was revised, and deformations for different forces and work points were presented. The results allow estimating the accuracy of technological process and improving it by the accurate planning of the milling.

While analyzing the modeling results, the several directions for prospective research activities were identified. They include accurate modeling of milling, improving stiffness of the manipulator for the working direction and compliance deformation compensation in the milling process.

## 7 ACKNOWLEDGMENTS

The work presented in this paper was partially funded by the Region “Pays de la Loire”, France and by the EU commission (project NEXT).

## 8 REFERENCES

[1] Brogardh, T., 2007, Present and future robot control development – an industrial perspective, Annual Reviews in Control 31/1: 69-79.

[2] Chanal, H., Duc, E., Ray, P., 2006, A study of the impact of machine tool structure on machining processes, International Journal of Machine Tools and Manufacture 46/2: 98-106.

[3] Merlet, J.-P., 2000, Parallel Robots, Kluwer Academic Publishers, Dordrecht.

[4] Koseki, Y., Tanikawa, T., Koyachi, N., Arai, T., 2000, Kinematic Analysis of Translational 3-DOF Micro Parallel Mechanism Using Matrix Method, Proc. of IROS 2000, W-AIV-7/1: 786-792

[5] Alici, G., Shirinzadeh, B., 2005, Enhanced stiffness modeling, identification and characterization for robot manipulators, IEEE Transactions on Robotics 21/4:554–564.

[6] Pashkevich, A., Chablat, D., Wenger, P., 2009, “Stiffness analysis of overconstrained parallel manipulators,” *Mechanism and Machine Theory*, 44:966-982

[7] Pashkevich A., Chablat D. Klimchik A. 2009. Enhanced stiffness modeling of serial manipulators with passive joints, Intech, Austria, In Press

[8] Pashkevich A.; Klimchik A.; Chablat D. & Wenger P., 2009, Stiffness analysis of multichain parallel robotic systems with loading, Journal of Automation, Mobile Robotics & Intelligent Systems, 3/3: 75-82

[9] Briot S., Pashkevich A. and Chablat D., 2010, Optimal Technology-Oriented Design of Parallel Robots for High-Speed Machining Applications, IEEE 2010 International Conference on Robotics and Automation, In Press

[10] Merlet J.-P., 2006, Jacobian, manipulability, condition number, and accuracy of parallel robots, Transaction of the ASME Journal of Mechanical Design, 128/1: 199–206.

[11] Merlet J.-P., 2006, Computing the worst case accuracy of a PKM over a workspace or a trajectory, The 5th Chemnitz Parallel Kinematics Seminar, Chemnitz, Germany: 83–96.

[12] El-Khasawneh, B.S. and Ferreira, P.M., 1999, Computation of Stiffness and Stiffness Bounds for Parallel Link Manipulators, International Journal of Machine Tools and Manufacture, 39/2: 321–342.

[13] Bouzgarrou, B.C., Fauroux, J.C., Gogu, G. and Heerah, Y., 2004, Rigidity Analysis of T3R1 Parallel Robot Uncoupled Kinematics, Proceedings of the 35th International Symposium on Robotics, Paris, March.

[14] Deblaise, D., Hernot, X. and Maurine, P., 2006, Systematic Analytical Method for PKM Stiffness Matrix Calculation, Proceedings of the IEEE International Conference on Robotics and Automation (ICRA), Orlando, Florida, May: 4213–4219.

[15] Ghali, A., Neville, A.M. and Brown, T.G., 2003, Structural Analysis: A Unified Classical and Matrix Approach, Spon Press, NY.

[16] Gosselin, C.M. and Zhang, D., 2002, Stiffness Analysis of Parallel Mechanisms Using a Lumped Model, International Journal of Robotics and Automation, 17/1: 17–27.

[17] Majou, F., Gosselin, C.M, Wenger, P. and Chablat, D., 2007, Parametric Stiffness Analysis of the Orthoglide, Mechanism and Machine Theory, 42/3: 296–311

[18] Pashkevich A., Klimchik A.; Chablat D. & Wenger P., 2009, Accuracy Improvement for Stiffness Modeling of Parallel Manipulators, Proceedings of 42nd CIRP Conference on Manufacturing Systems, June 2009

[19] Akin, J. E., 2005, Finite Element Analysis With Error Estimators: An Introduction to the FEM and Adaptive Error Analysis for Engineering Students Elsevier, Amsterdam.

[20] Chablat, D., Wenger, Ph., 2003, Architecture Optimization of a 3-DOF Parallel Mechanism for Machining Applications, the Orthoglide, IEEE Transactions on Robotics and Automation, 19/3: 403-410.